\documentclass[conference]{IEEEtran}
\IEEEoverridecommandlockouts
\usepackage{float}
\usepackage{amsmath,amssymb,amsfonts}
\usepackage{algorithmic}
\usepackage{graphicx}
\usepackage{textcomp}
\usepackage{xcolor}
\usepackage{caption}
\def\BibTeX{{\rm B\kern-.05em{\sc i\kern-.025em b}\kern-.08em
    T\kern-.1667em\lower.7ex\hbox{E}\kern-.125emX}}
\begin{document}

\title{Quadrant Segmentation VLM with Few-Shot Adaptation and OCT Learning-based Explainability Methods for Diabetic Retinopathy\\

}

\author{\IEEEauthorblockN{Shivum Telang}
\IEEEauthorblockA{\textit{Department of Biostatics} \\
\textit{North Allegheny Senior High School; University of Pittsburgh Rangos Research Center}\\
Pittsburgh, Pennsylvania \\
SHT283@pitt.edu}}

\maketitle

\begin{abstract}
Diabetic Retinopathy (DR) is a leading cause of vision loss worldwide, requiring early detection to preserve sight. Limited access to physicians often leaves DR undiagnosed. To address this, AI models utilize lesion segmentation for interpretability; however, manually annotating lesions is impractical for clinicians. Physicians require a model that explains the reasoning for classifications rather than just highlighting lesion locations. Furthermore, current models are one-dimensional, relying on a single imaging modality for explainability and achieving limited effectiveness. In contrast, a quantitative-detection system that identifies individual DR lesions in natural language would overcome these limitations, enabling diverse applications in screening, treatment, and research settings. To address this issue, this paper presents a novel multimodal explainability model utilizing a VLM with few-shot learning, which mimics an ophthalmologist's reasoning by analyzing lesion distributions within retinal quadrants for fundus images. The model generates paired Grad-CAM heatmaps, showcasing individual neuron weights across both OCT and fundus images, which visually highlight the regions contributing to DR severity classification. Using a dataset of 3,000 fundus images and 1,000 OCT images, this innovative methodology addresses key limitations in current DR diagnostics, offering a practical and comprehensive tool for improving patient outcomes.        
\end{abstract}


\section{Introduction}
Diabetic retinopathy (DR) is an aggressive retinal disease that is one of the leading causes of vision loss
and blindness in the world. Just in the United States, 9.60 million people have DR. Of those 9.6 million
people, 30 percent have Proliferative Diabetic Retinopathy constituting retinal detachment or vision loss \cite{Zhang2010}. Treatment of DR usually consists of photocoagulation using laser treatment to slow the
leakage of blood to stop the abnormal growth of blood vessels in the retina or Anti-VEGF injections
releasing a drug in the vitreous to contain the growth of the disease \cite{Mansour2020}. For
ophthalmologists to treat DR, the severity level must first be classified before starting treatment. DR is
classified based on two types of retinal scans: fundus and OCT. A fundus scan shows a picture of the
retina including lesions, blood vessels, optic nerve, etc. On the other hand, the OCT scan shows the
structure of the retina, including intraretinal structures, fluids, and layers \cite{YannuzziND}.

To improve visual interpretability, current state-of-the-art models integrate lesion segmentation to individually annotate each lesion on a fundus image. Segmentation models use a U-Net encoder-decoder architecture and have attempted to be effective in identifying and analyzing lesions in retinal scans \cite{Alharbi2023}. A U-Net uses a contracting path (encoder) which reduces the spatial resolution while increasing depth to extract key image features and an expansive path (decoder) which reconstructs the image while preserving spatial information. While this method proves useful for disease detection, it has not yet been widely implemented in clinical practice, due to its lack of explainability in natural language \cite{Kamal2024}.

Furthermore, ophthalmologists rely on both fundus and optical coherence tomography (OCT) images for severity classification, whereas most AI models use only a single imaging modality. This limitation further makes it difficult for physicians to use most AI models signifying the need for an explainable diagnostic model.

\section{Related Works}

To fix the “black box” interpretability problem, previous models have used SHAP for explainable classification on OCT images by generating an overlay map to show regions contributing positively to the respective classification. However, the SHAP algorithm has issues of assuming features (pixels or regions) as independently removed and affects the spatial relation of the image \cite{Feng2021}. Past models have also tried using LIME values for greater explainability, but this has also faced problems of reliance on super pixels making clustering arbitrary and regions not correctly aligning with boundaries \cite{Gabbay2021}. To overcome these challenges in model explainability, this paper utilizes Grad-CAM which generates a heatmap based on the activations of the neurons in the last convolutional layer of the model to account for important features or regions \cite{Selvaraju2016}.
For greater interpretability of fundus images, current models use segmented U-Nets to extract lesions and features contributing to the classification \cite{Alharbi2023}. These segmentation models cannot reach mean pixel accuracy of greater than 85 percent and still lack trust in segmentation because of the inability of explaining the results in natural language to aid ophthalmologists in their clinical decision-making process \cite{Kamal2024}. Therefore, to tackle this interpretability problem for fundus images, this paper utilizes a VLM model with few-shot learning to focus on finding the lesions and type through texture channels and color channels in each quadrant and displayed in natural language to the clinician \cite{Muragappan2022}.

\section{Methodology}

This paper proposes a novel state-of-the-art approach for explainable methods of DR. Below is a detailed explanation of the methodology for each component of the model framework:

\subsection{Data Curation}

In the study, we utilized two datasets. For the two datasets, we utilized 3,500 fundus images collected and 1,000 OCT images collected from public datasets \cite{Naren2021,Cukierski2015}. For each retinal image was information about the date of the scan, the patient’s age, and the patient’s level of diabetes. The figure below shows example scans for fundus and OCT images:

\begin{center}
\includegraphics[width=\linewidth]{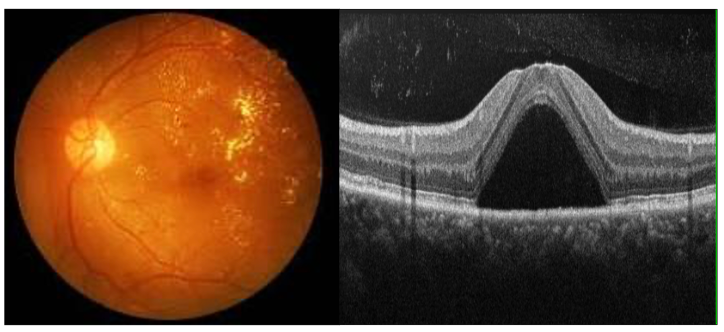}
\captionof{figure}{The image to the left shows a fundus scan (picture of retina) and the image to the right shows an OCT scan (structure of retina)}
\label{fig:gradcam}
\end{center}

\subsection{Grad-CAM Explainability Framework}

To explain the classification results to an ophthalmologist, we add a unique Grad-CAM \cite{Selvaraju2016} based heatmap onto the input OCT images and a quadrature segmentation model for fundus scans for desirable applicability.

To generate the heat maps, we first take the last convolutional layer of the previous multimodal network and implement a Grad-CAM heatmap architecture on the OCT scan \cite{LinJhang2021}. The models use an activation map based on spatial indices from the input’s feature map. Then the network determines the feature maps that are important for class $c$ by computing the gradients of the class score $S_c$ with respect to feature maps $A^k$. The operation determines how much the activation map $A^k$ contributes to the decisions for class $c$. Then the class-activation map is calculated through a weighted sum of the feature maps. Based on this function, the model generates a low-resolution heat map, $L_{\text{GradCAM}}^c$ \cite{Selvaraju2017}. To match the sample input image size, we use bilinear image interpolation to upsample the input. This operation can be modeled by the equation below:

\[
\tilde{L}_{\text{GradCAM}}^c = \text{Upsample}(L_{\text{GradCAM}}^c, \text{size} = (H_0, W_0))
\]

In the equation above, $H_0$ and $W_0$ represent the original input image dimensions. After upsampling the heatmap, it is overlayed onto the input image to visualize the regions contributing to the DR classifications. The overlay function can be modeled as such:

\[
\text{Heatmap} = \lambda \tilde{L}_{\text{GradCAM}}^c + (1 - \lambda)X
\]

In this operation, the variable $\lambda$ controls the blending of the heatmap into the original image. The sample input image was originally used as fundus and specified for then OCT. The full Grad-CAM architecture is shown in \textbf{Figure~\ref{fig:gradcam}}.

\begin{center} % <--- Use the center environment to confine centering
\includegraphics[width=\linewidth]{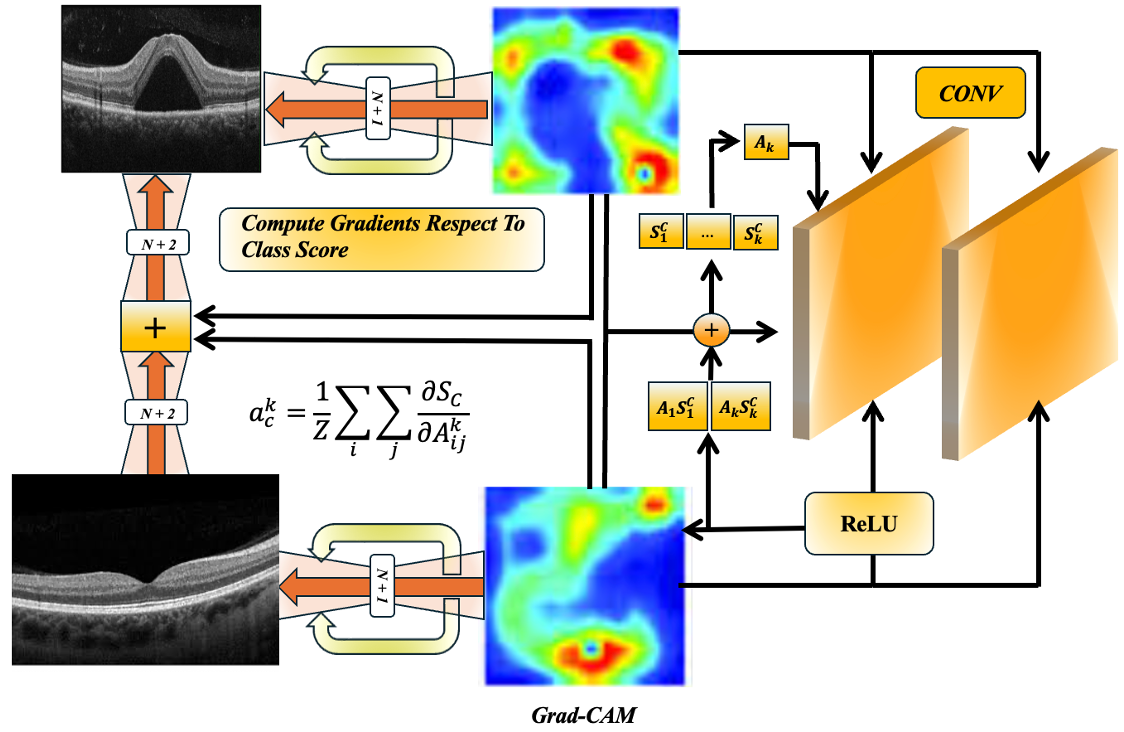}
\captionof{figure}{Grad-CAM Heatmap Architecture }
\label{fig:gradcam}
\end{center}
\vspace{-10pt}
\subsection{Quadrant Segmentation VLM with Few-Shot Learning}

The segmentation framework targets key retinal lesions within fundus images, including microaneurysms, dot-blot hemorrhages, hard exudates, and cotton-wool spots. A VLM trained via few-shot learning processes fundus images divided into four quadrants on a 2D plane centered at (0,0) \cite{Muragappan2022}. A region proposal network (RPN) employing a Faster R-CNN identifies candidate lesion regions, which are then passed downstream for classification \cite{Albahli2020}. The RPN outputs feature maps with bounding box proposals $B_i$ and objectness scores $p_i$optimized through cross-entropy loss for objectness and smooth L1 loss for box regression as defined in the equation below \cite{Brazil2019}.

\[
L_{RPN} = \frac{1}{N_{pos}} \sum_i p_i^{gt} * SmoothL1(t_i, t_i^{gt})
\]

The loss then outputs all the lesions in each specified quadrant the filter is being applied to. This information is then combined into VLM treating both visual and text inputs in its joint embedding space. Based on the few-shot learning characterization of lesions, this can then be converted to a natural language explanation of the number and type of lesions in each quadrant to create a comprehensive explainability framework.  The whole architecture is outlined in \textbf{Figure~\ref{fig:VLM}} below.

\begin{figure*}
\begin{center}
 \includegraphics[width=\linewidth]{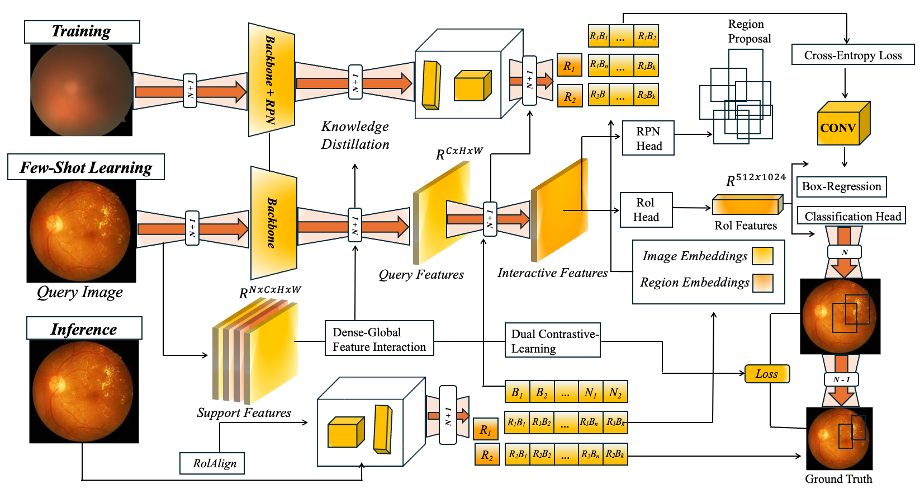}
 \caption{VLM with Few-Shot Learning Explainability Filter }
 \label{fig:VLM}
\end{center}
\end{figure*}
\vspace{10pt}
\section{Results}

\noindent The table shows the VLM results for the different lesions. 

\begin{table}[H]
\renewcommand{\arraystretch}{1.2}
\centering
\caption{Detection performance of DR lesion types.}
\begin{tabular}{lcc}
\hline
\textbf{DR lesions} & \textbf{Count in manual results} & \textbf{Sensitivity} \\
\hline
Microaneurysm & 797 & 96.4\% (768/797) \\
Hemorrhage & 1623 & 99.6\% (1617/1623) \\
Hard exudates & 2319 & 99.5\% (2308/2319) \\
Soft exudates & 86 & 94.2\% (81/86) \\
\hline
\end{tabular}
\label{tab:dr_lesion_performance}
\end{table}

\noindent Below \textbf{Figure~\ref{fig:VLMResults}} shows a sample response from the VLM with natural language: explaining the number and type of lesions within each quadrant. 

\begin{center} % <--- Use the center environment to confine centering
\includegraphics[width=\linewidth]{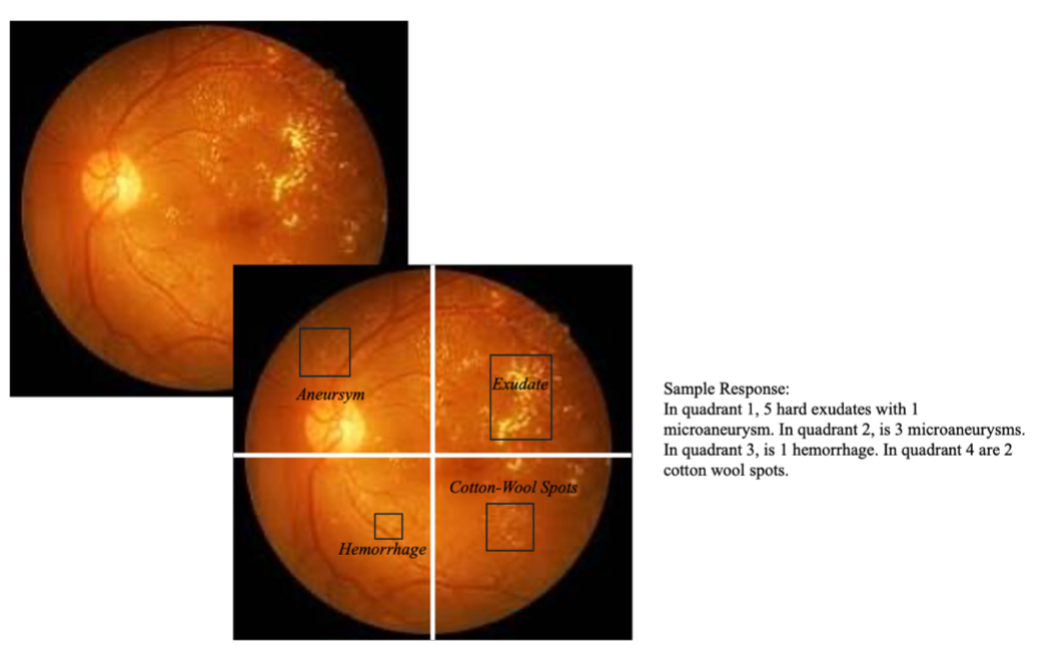}
\captionof{figure}{Fundus Image Quadrant-based Lesion Extraction Response}
\label{fig:VLMResults}
\end{center}
\vspace{-10pt}

\noindent \textbf{Figure~\ref{fig:GradCAMResults}} below shows the Grad-CAM heatmap overlay representing the regions contributing most the classificatiion for the various classes on OCT scans. 

\begin{center} % <--- Use the center environment to confine centering
\includegraphics[width=\linewidth]{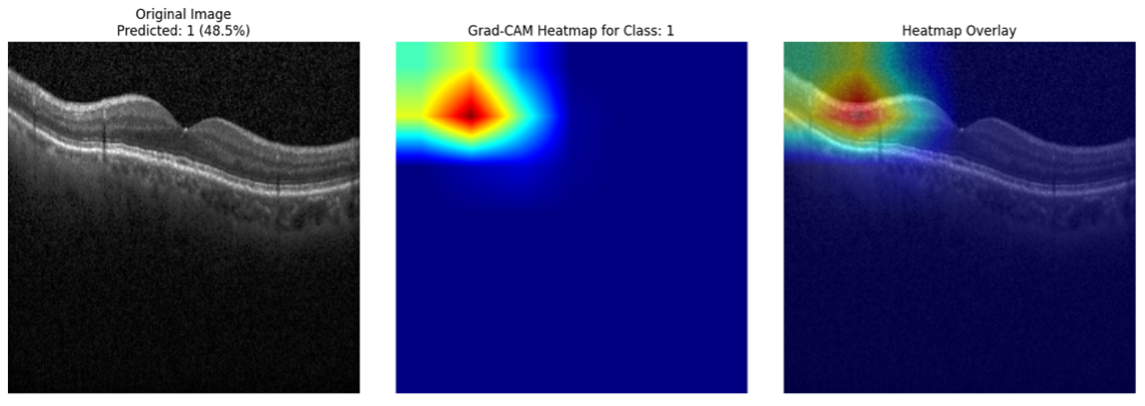}
\captionof{figure}{Grad-CAM Heatmap overlay for different classes}
\label{fig:GradCAMResults}
\end{center}

\section{Conclusion and Future Work}

This study introduces a unique quadrant segmentation framework for fundus images integrated with Grad-CAM explainability to enhance lesion-level interpretation in screening and treatment for DR. By partitioning fundus images into quadrants and applying a region proposal network, the system achieved precise localization of key retinal lesions, including microaneurysms, hemorrhages, and exudates. The Grad-CAM visualization provided interpretable heatmaps that highlight the regions that contribute most to the severity level and preventing the 'black-box' interpretability problem. 

The results showed high sensitivity in the various lesion categories, confirming the effectiveness of the VLM few-shot learning based explainability in aiding clinical decision-making. This framework bridges the gap between deep learning performance and clinical interpretability, offering a practical method for transparent screening.

 Future work for RetGEN can include using longitudinal patient data for temporal analysis of disease progression, developing a federated learning model for decentralized model training access across various clinical settings, applying RetGEN to other retinal conditions, and testing other clinical explainability methods such as diffusion transformers.

 \bibliographystyle{ieeetr}
    \bibliography{references}

\end{document}